\documentclass[pdflatex,sn-mathphys-num]{sn-jnl}

\usepackage{graphicx}%
\usepackage{multirow}%
\usepackage{amsmath,amssymb,amsfonts}%
\usepackage{amsthm}%
\usepackage{mathrsfs}%
\usepackage[title]{appendix}%
\usepackage{xcolor}%
\usepackage{textcomp}%
\usepackage{manyfoot}%
\usepackage{booktabs}%
\usepackage{algorithm}%
\usepackage{algorithmicx}%
\usepackage{algpseudocode}%
\usepackage{listings}%
\usepackage[inline]{trackchanges}

\theoremstyle{thmstyleone}%
\theoremstyle{thmstyletwo}%

\theoremstyle{thmstylethree}%

\begin{document}

\title[Urban Greening Choices Matter]{%
Physiologically Active Vegetation Reverses Its Cooling Effect in Humid Urban Climates}

\author[1]{\fnm{Angana} \sur{ Borah}}\email{borah\_angana@iitgn.ac.in}

\author[2]{\fnm{Adrija} \sur{ Datta}}\email{dattaadrija@iitgn.ac.in}

\author[1]{\fnm{Ashish } \sur{S. Kumar}}\email{ashishkumar@iitgn.ac.in}

\author[1,3]{\fnm{Raviraj} \sur{Dave}}\email{r.dave@northeastern.edu}

\author*[1,2,4]{\fnm{Udit} \sur{Bhatia}}\email{bhatia.u@iitgn.ac.in}

\affil[1]{\orgdiv{Department of Civil Engineering}, \orgname{Indian Institute of  Technology Gandhinagar}, \orgaddress{\street{Palaj}, \city{Gandhinagar}, \postcode{382355}, \state{Gujarat}, \country{India}}}

\affil[2]{\orgdiv{Department of Earth Sciences}, \orgname{Indian Institute of  Technology Gandhinagar}, \orgaddress{\street{Palaj}, \city{Gandhinagar}, \postcode{382355}, \state{Gujarat}, \country{India}}}

\affil[3]{\orgdiv{Sustainability and Data Sciences Laboratory, Department of Civil and Environmental Engineering}, \orgname{Northeastern University}, \orgaddress{ \city{Boston}, \state{MA}, \country{USA}}}

\affil[4]{\orgdiv{Department of Computer Sciences and Engineering}, \orgname{Indian Institute of  Technology Gandhinagar}, \orgaddress{\street{Palaj}, \city{Gandhinagar}, \postcode{382355}, \state{Gujarat}, \country{India}}}

\abstract{
Efforts to green cities for cooling are succeeding unevenly because the same vegetation that cools surfaces can also intensify how hot the air feels. Previous studies have identified humid heat as a growing urban hazard, yet how physiologically active vegetation governs this trade-off between cooling and moisture accumulation remains poorly understood, leaving mitigation policy and design largely unguided. Here we quantify how vegetation structure and function influence the Heat Index (HI)—a combined measure of temperature and humidity—in 138 Indian cities spanning tropical savanna, semi-arid steppe, and humid subtropical climates, and across dense urban cores and semi-urban rings. Using an extreme-aware, one-kilometre reconstruction of HI and an interpretable machine-learning framework that integrates SHapley Additive Explanations (SHAP) and Accumulated Local Effects (ALE), we isolate vegetation–climate interactions. Cooling generally strengthens for EVI~$\geq$~0.4 and LAI~$\geq$~0.05, but joint-high regimes begin to reverse toward warming when EVI~$\geq$~0.5, LAI~$\geq$~0.2, and fPAR~$\geq$~0.5, with an earlier onset for fPAR~$\geq$~0.25 in humid, dense cores. In such environments, highly physiologically active vegetation elevates near-surface humidity faster than it removes heat, reversing its cooling effect and amplifying perceived heat stress. These findings establish the climatic limits of vegetation-driven cooling and provide quantitative thresholds for climate-specific greening strategies that promote equitable and heat-resilient cities.}

\maketitle


\section*{Introduction}\label{sec1}
Cities worldwide are warming faster than their rural surroundings \cite{liu2022surface}, exposing growing populations to dangerous heat stress that threatens health, productivity, and livability \cite{mandvikar2024evaluating, vanos2023physiological}. The combination of urbanization and global warming has intensified air temperatures \cite{zhou2022urbanization}, while local moisture and limited ventilation often keep humidity high\cite{huang2023urban}, amplifying the perceived heat beyond what dry bulb temperature alone indicates and reshaping the nature of urban heat exposure. This combined burden of temperature and moisture, measured through indices such as the Heat Index (HI)\cite{willett2012exceedance}, now defines the lived experience of climate risk in cities. Prolonged exposure to high HI has been linked to increased mortality, illness, and economic losses \cite{amnuaylojaroen2024projections}, particularly in South and Southeast Asia \cite{jacobs2019patterns, saeed2021deadly}, where dense populations and humid monsoonal climates converge \cite{yin2023urban, sethi2024urbanization}. However, the burden is uneven: poorer neighborhoods and outdoor workers face the greatest exposure, making urban heat a clear dimension of climate inequality.

\begin{figure}[H]
    \centering
    \includegraphics[width=0.9\textwidth]{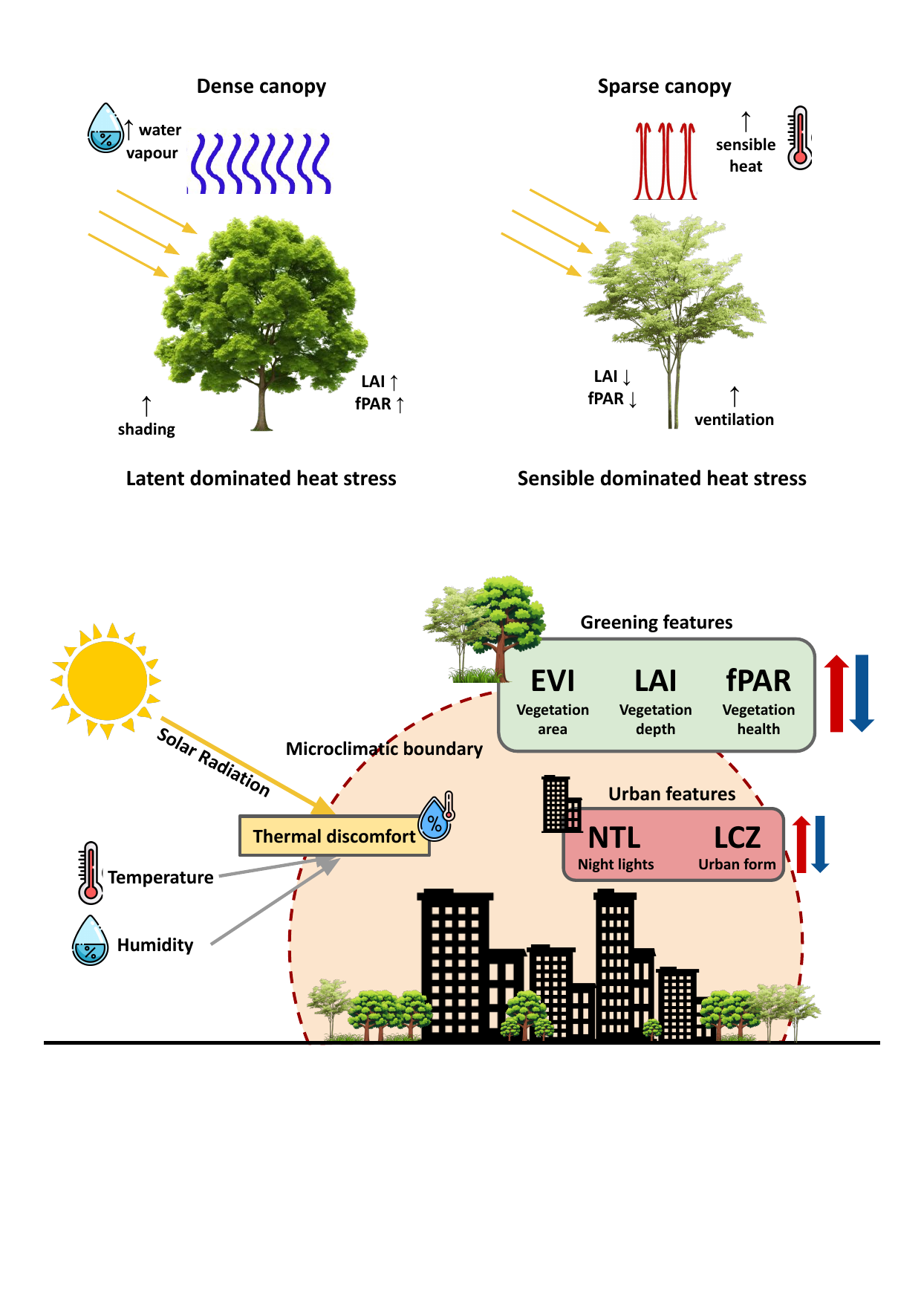}
    \caption{
    \textbf{Conceptual framework linking canopy structure, greening features, and urban form to thermal discomfort.}
    The upper panel illustrates how canopy density regulates heat stress mechanisms: dense canopies enhance shading and evapotranspiration through higher Leaf Area Index (LAI) and Fraction of Absorbed Photosynthetically Active Radiation (fPAR), leading to latent-dominated heat stress; whereas sparse canopies promote ventilation but increase sensible heat flux due to lower LAI and fPAR, resulting in sensible-dominated heat stress. The lower panel conceptualizes how these processes interact with urban morphology within a microclimatic boundary. Solar radiation, air temperature, and humidity define the ambient thermal environment, while greening features such as, Enhanced Vegetation Index (EVI; vegetation area), LAI (canopy depth), and fPAR (vegetation health) and urban features such as night-time lights (NTL; built intensity) and Local Climate Zones (LCZ; urban form), jointly regulate local heat exchange. Their combined influence determines the degree of thermal discomfort experienced within urban neighborhoods.
    }
    \label{fig:1}
\end{figure}

Urban greening is widely promoted as a natural solution to mitigate heat \cite{li2025global, wong2021greenery}, yet its performance varies markedly between climates and city forms \cite{li2024cooling}. On hot days, the air near the vegetation can be cooler, but still feel hotter if the added moisture is trapped by the limited airflow. This paradox means that greening can help or harm depending on the background climate, local ventilation, and which facet of vegetation is modified. Policies often assume that any additional greenery will reduce heat, but the reported outcomes differ depending on the neighborhood context, the formulation of the model, and the scale of intervention. In practice, cities make choices about trees, parks and reflective surfaces with incomplete guidance on when shading and roughness dominate and when transpiration-driven humidity erodes comfort gains. Understanding when and where greening improves human comfort is therefore essential for designing effective and equitable adaptation strategies (Fig.~\ref{fig:1}).

Vegetation moderates urban heat through two primary pathways: shading \cite{rahman2021comparative}, which reduces the absorption of solar radiation by built surfaces, and evapotranspiration \cite{zhang2013effects}, which converts sensible heat into latent cooling. These processes typically lower local temperature and alter aerodynamic roughness and the balance of long-wave radiation \cite{yang2024regulation, kumar2024urban}. However, in humid or poorly ventilated environments, the same evapotranspiration adds moisture to the air faster than it can be dispersed, reducing the net cooling benefit or even intensifying perceived heat \cite{chen2025contrasting, li2024cooling}. This tension between radiative and latent processes, which is fundamental to how people experience thermal comfort, is poorly understood in all climatic and morphological contexts, leaving mitigation design largely unguided. Most studies evaluate the effects of vegetation using the land surface or air temperature, rarely accounting for humidity or its contribution to discomfort \cite{du2025exacerbated, luo2021increasing}. Clarifying how shading and evapotranspiration interact with local moisture and urban form is therefore central to designing effective and equitable cooling strategies in the world's fastest-urbanized regions.

Previous studies have extensively examined the phenomenon of the Urban Heat Island (UHI) and the cooling potential of vegetation using land-surface or air temperature as primary indicators \cite{hao2023urbanization, yang2024regulation}. Although such approaches have advanced understanding of surface thermal regimes, they overlook the critical role of humidity in shaping human thermal perception and comfort \cite{luo2021increasing, du2025exacerbated}. Indices that explicitly integrate temperature and moisture, such as the Heat Index (HI) or Universal Thermal Climate Index (UTCI), remain underrepresented in large-scale urban analyses, particularly in tropical and monsoonal regions where humidity dominates heat stress \cite{yin2023urban}. In addition, vegetation and built-form variables are commonly treated as independent predictors, providing limited information on their coupled and nonlinear effects on thermal discomfort \cite{chen2025contrasting}. Consequently, despite the growing recognition that vegetation can both alleviate and amplify heat stress, we still lack a mechanistic understanding of how shading, evapotranspiration, and humidity interact across diverse climates and densities. This absence restricts our ability to generalize comfort outcomes or develop actionable climate-specific adaptation strategies. Recent analysis has shown that in humid climates, urban air can become more oppressive than its rural surroundings - not due to higher temperatures alone, but because of the accumulation of moisture and reduced atmospheric mixing that limit the dissipation of convective heat \cite{zhang2023increased}. This shift from a dry to a moist urban island challenges the assumption that greening universally improves comfort. Overall, these studies did not explicitly model vegetation structure or function, nor did they resolve the climate-specific thresholds at which greening transitions from beneficial to detrimental. 

Here, we address this gap using a unified, data-driven framework that links vegetation structure and function with urban morphology to explain humidity-adjusted heat stress across India. We analyze 138 cities spanning tropical savanna (Aw), semi-arid steppe (BSh), and humid-subtropical (Cwa) climates \cite{beck2018present}. Vegetation characteristics are represented by the Leaf Area Index (LAI), Enhanced Vegetation Index (EVI), and Fraction of Absorbed Photosynthetically Active Radiation (fPAR), which together describe canopy depth, surface greenness, and photosynthetic activity \cite{huete2002overview, myneni2002global}. Urban form and intensity are captured by night-time lights (NTL) and Local Climate Zones (LCZ) \cite{zheng2023nighttime, stewart2012local}, which, together with vegetation indices, form the predictor framework illustrated in Fig.~\ref{fig:1}. We develop an extreme-aware, high-resolution reconstruction of the Heat Index using reanalysis temperature and humidity downscaled to 1~km, and apply an attribution framework combining SHapley Additive exPlanations (SHAP) with Accumulated Local Effects (ALE) to quantify both individual and joint influences. Specifically, we ask three questions: (1) which greening and urban-form features most strongly explain thermal discomfort; (2) how these features interact—i.e., which pairs exert the largest joint influence on HI—identified via SHAP interaction analysis and visualized with two-dimensional ALE; and (3) how much HI (in $^{\circ}\mathrm{C}$) changes for specific changes in each key feature and for the top feature pairs, and how much greening is required to achieve a target reduction in HI to guide mitigation. This approach directly models the humidity-adjusted discomfort experienced by people rather than surface temperature alone, linking biophysical canopy traits with human-perceived heat. By moving beyond temperature-based metrics, this study converts descriptive evidence of urban cooling into quantitative attribution of vegetation–climate interactions, providing a physically constrained and policy-relevant foundation for designing equitable and heat-resilient cities.

\section*{Results}\label{sec2}

\subsection*{Spatial and temporal heterogeneity of Heat Index}

\begin{figure}[H]
    \centering
    \includegraphics[width=0.9\textwidth]{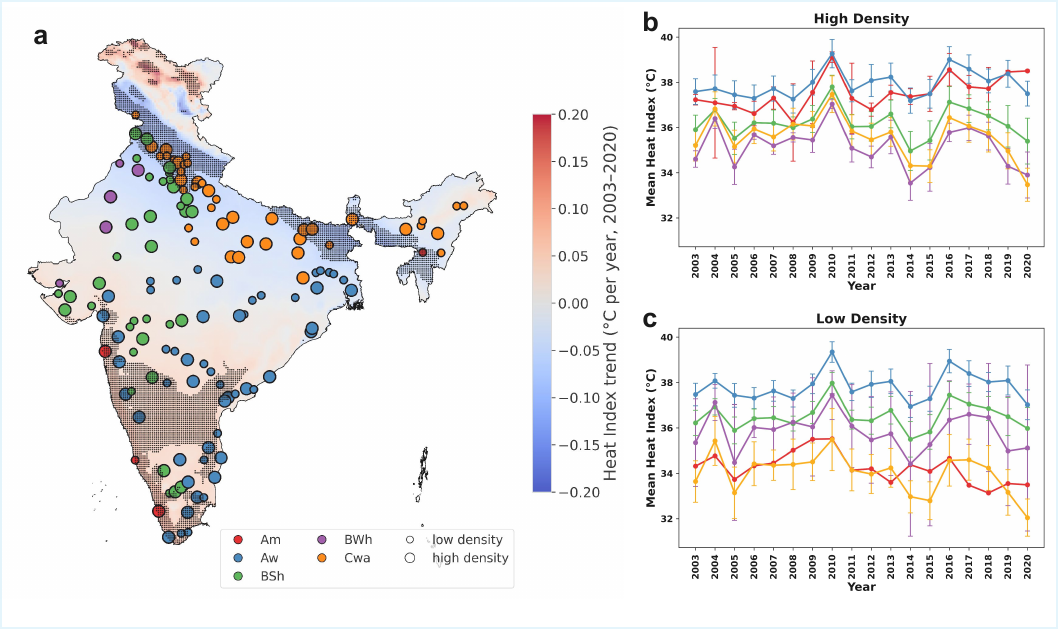}
    \caption{\textbf{Spatial and temporal patterns of Heat Index (HI).}
    (a) City-level linear trends in HI (°C\,yr$^{-1}$; 2003–2020) across 138 Indian cities, grouped by Köppen–Geiger class and population density (high/low). Stippling marks $p<0.05$. (b,c) Interannual variability of mean HI for high- and low-density cohorts, respectively; vertical bars denote $\pm1$\,SD across cities in each class.}
    \label{fig:2}
\end{figure}

Thermal discomfort, expressed as the Heat Index (HI), exhibits pronounced spatial and temporal variability across Indian cities (Fig.~\ref{fig:2}). Linear trends for 2003–2020 reveal coherent warming clusters ($p<0.05$) concentrated in the arid and semi-arid northwest and along the central peninsular corridor, while scattered cooling pockets appear in high-elevation and forested regions, including the Himalayan foothills and parts of the northeast (Fig.~\ref{fig:2}a). Stratification by Köppen–Geiger climate shows that warm-temperate, humid-subtropical cities (Cwa) generally display positive HI trends, whereas tropical (Aw) and semi-arid (BSh) cities show mixed patterns that coincide with variations in local moisture and land cover. Interannual variability further distinguishes dense and low-density cohorts (Fig.~\ref{fig:2}b,c): densely constructed cores sustain persistently high HI baselines with limited fluctuation, while semi-urban rings exhibit larger year-to-year variability (Fig.~\ref{fig:2}b,c), consistent with stronger exposure to meteorological variability and land-use dynamics. Collectively, these spatial and temporal contrasts outline the background heterogeneity of urban heat exposure and frame subsequent analyses that attribute HI variability to vegetation and urban-form features using SHapley Additive exPlanations (SHAP) and Accumulated Local Effects (ALE) (See Methods).

\subsection*{Vegetation and urban-form controls on Heat Index}

\begin{figure}[H]
    \centering
    \includegraphics[width=0.9\textwidth]{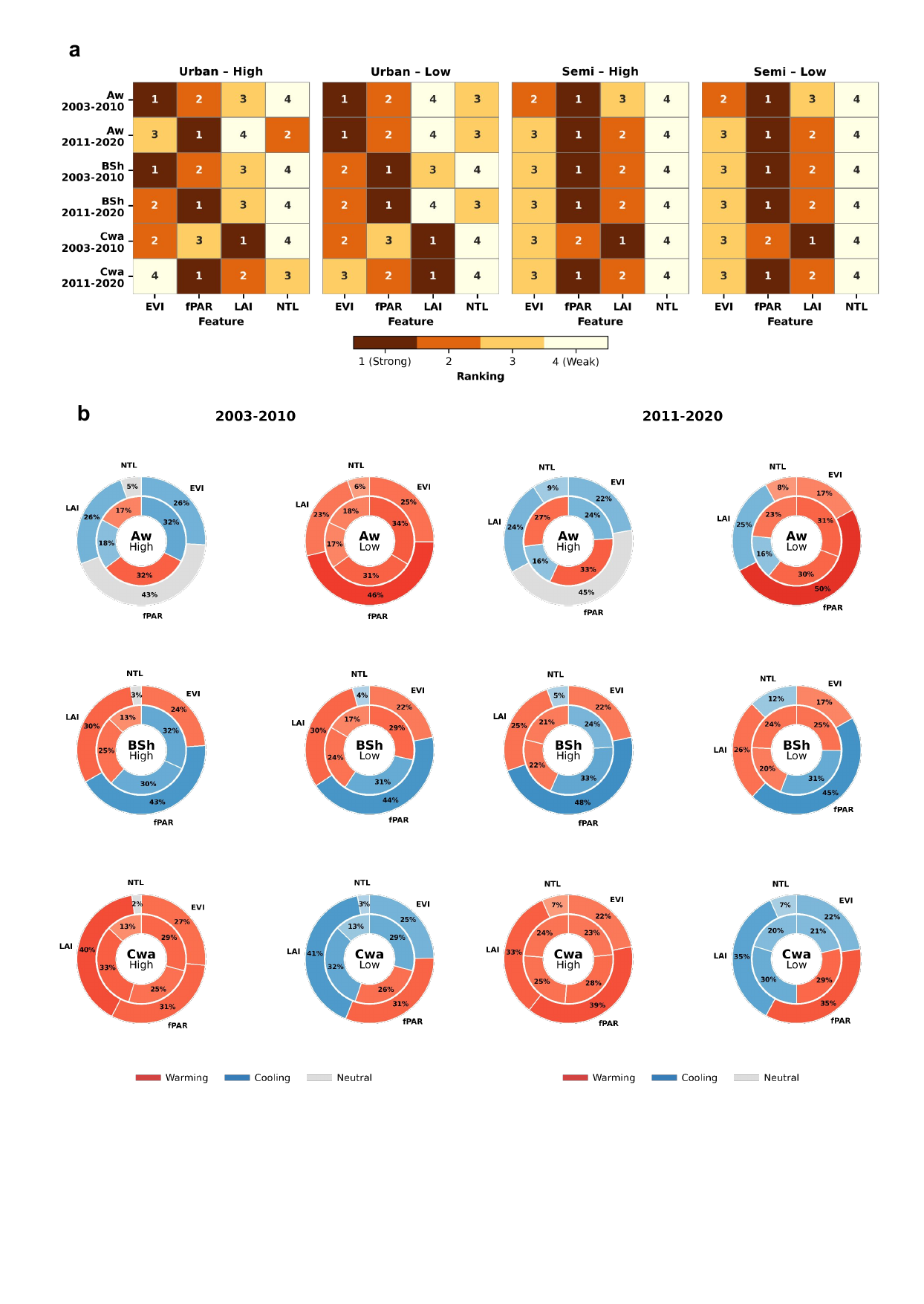}
    \caption{\textbf{Relative importance and direction of greening and urban-form features on HI.}
    (a) Feature importance derived from SHapley Additive exPlanations (SHAP) for EVI, fPAR, LAI, and NTL across Aw, BSh, and Cwa classes and urbanization strata (Urban–High/Low, Semi–High/Low) for 2003–2010 and 2011–2020. Darker shading indicates stronger explanatory power (rank\,=\,1).
    (b) Decadal shifts in the direction and strength of associations shown as paired donut charts; inner rings represent urban cores and outer rings semi-urban zones.}
    \label{fig:3}
\end{figure}

Across climates and density strata, vegetation features account for most of the observed spatial variability in Heat Index (HI), consistently outranking proxies of urban intensity (Fig.~\ref{fig:3}). Among vegetation metrics, the Fraction of Absorbed Photosynthetically Active Radiation (fPAR) is the strongest predictor in semi-arid steppe (BSh) cities, followed by the Leaf Area Index (LAI), whereas Night-Time Lights (NTL) contribute least across all groups. In humid-subtropical (Cwa) cities, LAI frequently leads with fPAR second, while in tropical savanna (Aw) settings, dominance alternates over time: the Enhanced Vegetation Index (EVI) dominates dense cores during 2003–2010, but fPAR becomes dominant in 2011–2020. These rankings are derived from SHapley Additive exPlanations (SHAP), which attribute each predictor’s relative contribution to HI variance.

The direction of association between vegetation features and HI varies strongly with background climate and urban density. In high-density cores of Aw, EVI and LAI are associated with lower HI values (approximately 18–32\%), but these relationships reverse to moderate positive associations (17–34\%) in low-density cores, where increased evapotranspiration coincides with elevated near-surface humidity. In contrast, the cooling effect of EVI and LAI increases from urban core to semi-urban in both decades (Fig.~\ref{fig:3}b). fPAR in Aw core shows more persistent positive associations towards warming (30–50\%). In BSh, fPAR is consistently linked to lower HI across densities (31–48\%), while EVI and LAI are linked to warming spatiotemporally. In subtropical humid Cwa high-density cities, all three greening variables correspond to higher HI in both urban cores and semi-urban strata, whereas in Cwa low-density cities, LAI and EVI show negative associations (25–42\%) and fPAR remains positive (26–35\%) across different densities. Corresponding decadal bar plots of mean signed $|\mathrm{SHAP}|$ contributions for all Köppen-density groups are provided in the (Supplementary \textcolor{blue}{Fig. S1}). These contrasts indicate that vegetation structure and function strongly modulate HI variability, but their net influence reverses with background moisture and urban morphology—a nonlinearity explored in the next subsection.

\subsection*{Joint effects of vegetation attributes on Heat Index}

\begin{figure}[H]
    \centering
    \includegraphics[width=0.9\textwidth]{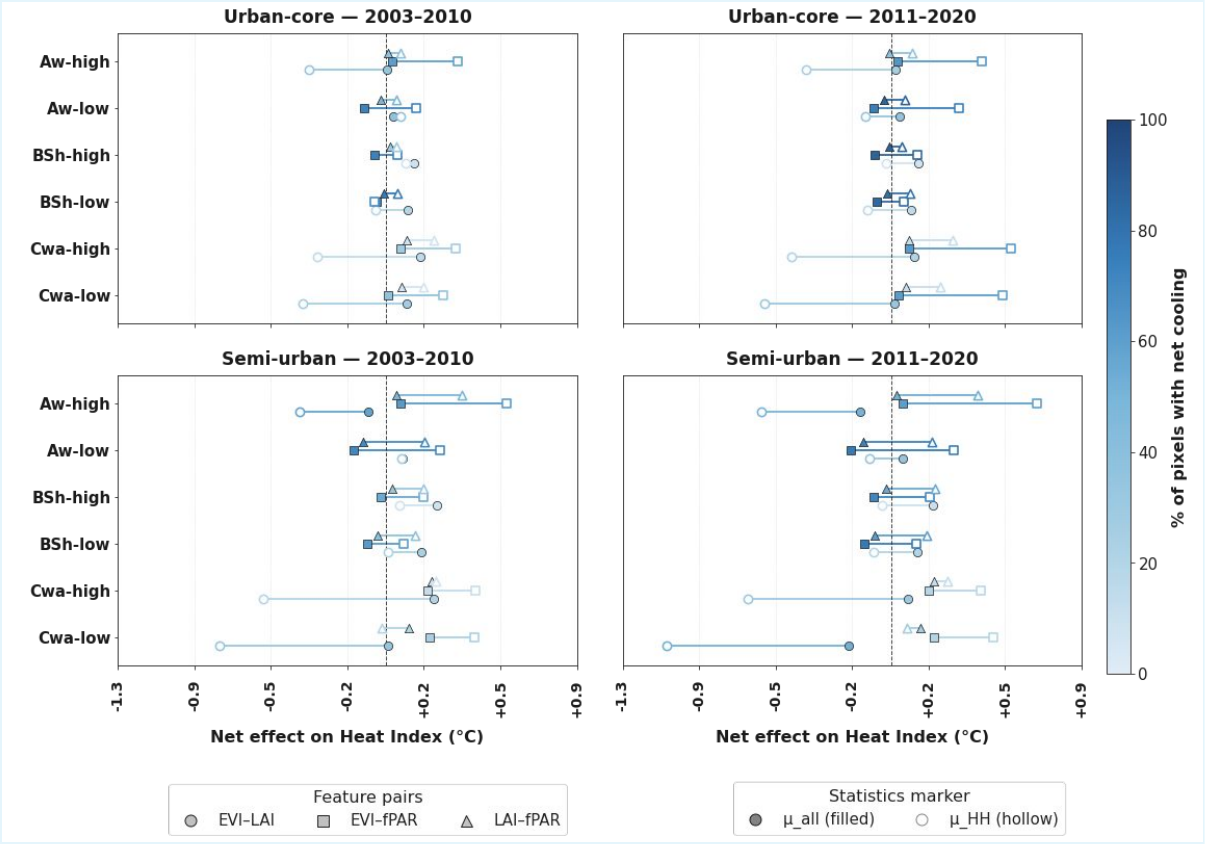}
    \caption{\textbf{Joint effects of vegetation attributes on Heat Index (HI).}
    Filled markers show $\mu_{\mathrm{all}}$ (domain mean) and hollow markers $\mu_{\mathrm{HH}}$ (joint-high mean) for Urban-core (top) and Semi-urban (bottom) zones during 2003–2010 (left) and 2011–2020 (right). Circles\,=\,EVI–LAI, squares\,=\,EVI–fPAR, triangles\,=\,LAI–fPAR. Horizontal bars show $\pm1$\,SD. Color intensity indicates cooling coverage (\% of pixels with net SHAP $<0$); negative values denote associations with lower HI, positive with higher HI.}
    \label{fig:4}
\end{figure}

The preceding analysis showed that vegetation features are key correlates of Heat Index (HI) variability, but their relative influence shifts across climates and densities, suggesting potential interactions between canopy structure and function. To examine these compound effects, we evaluated how vegetation attributes act jointly rather than independently. Pairwise SHapley Additive Explanation (SHAP) maps summarize the combined associations of two features with HI as the net deviation from baseline, expressed through three descriptive metrics: the domain-wide mean ($\mu_{\mathrm{all}}$), the joint-high mean ($\mu_{\mathrm{HH}}$), and the cooling coverage (percentage of pixels with net SHAP $<0$) across climate zones and density strata (Fig.~\ref{fig:4}). \\
For both tropical (Aw) and subtropical (Cwa) cities, and in both density classes, \(\mathrm{EVI}\)--\(\mathrm{fPAR}\) combinations lead to stronger warming (\(\approx +0.2\) to \(+0.7~^{\circ}\mathrm{C}\)) at joint high values of both features over the two decades. However, the warming is more pronounced in high-density cities than in low-density ones because of humidity trapping within built-up areas. In contrast, \(\mathrm{EVI}\)--\(\mathrm{LAI}\) combinations are associated with cooling across both Aw and Cwa regions, with notably stronger cooling (\(\approx -0.8\) to \(-1.0~^{\circ}\mathrm{C}\)) observed in semi-urban areas during the later decade (Supplementary \textcolor{blue}{Fig. S2}).\\
In Aw cores, structural greenness pairs (EVI–LAI) are linked to cooling at joint-high values($\mu_{\mathrm{HH}} \approx -0.38$ to $-0.51$\,°C) for high density cores (Fig.~\ref{fig:4}), whereas greening area-health pairs (EVI–fPAR) correspond to higher HI ($\mu_{\mathrm{HH}} \approx +0.33$ to $+0.69$\,°C), (Fig.~\ref{fig:4}) indicating a humidity-sensitive trade-off when vegetation activity increases under limited ventilation in already humid atmosphere. In BSh high-density urban cores, where evaporative demand is strong, fPAR-involving pairs are broadly associated with lower HI ($\mu_{\mathrm{all}}\le0$, cooling coverage 74-87\%) whereas in semi-urban rings the colling effect is less with ($\mu_{\mathrm{all}}\le0$, cooling coverage 55-72\%) (Fig.~\ref{fig:4}). In humid Cwa high-density cores, EVI-fPAR combinations are mainly associated with higher HI ($\mu_{\mathrm{HH}} \approx +0.28$ (2003-2010) to $+0.57$\,°C (2011-2020)) explaining the more warming pattern associated with greening (Fig.~\ref{fig:4}). However, for semi-urban region, such decadal change is not so prominent. In contrast, EVI–LAI shows modest cooling at joint-high values ($\mu_{\mathrm{HH}} \approx -0.33$ (2003-2010) to $-0.47$\,°C (2011-2020)) with the same pattern in low-density cities (Fig.~\ref{fig:4}).\\
The spatiotemporal opposite pattern between \(\mu_{\mathrm{all}}\) and \(\mu_{\mathrm{HH}}\) arise from a threshold-dependent reversal in the thermal influence of vegetation (Fig.~\ref{fig:4}). In LAI-involved combinations of humid-subtropical (Cwa) cities, especially within high-density cores across both decades, low vegetation cover (\(\mathrm{EVI}<0.5\), \(\mathrm{LAI}<0.1\)) enhances sensible heat flux and leads to warming (Supplementary \textcolor{blue}{Fig. S2}), but beyond this threshold, increased canopy density promotes shading and evapotranspiration, resulting in cooling. 
This reverse sweep reflects a transition from radiative trapping to latent heat dominance and remains consistent across all density levels and subtropical zones. In contrast, for \(\mathrm{EVI}\)-\(\mathrm{fPAR}\) combinations in humid-subtropical (Cwa) regions, the relationship reverses: high humidity limits latent cooling, so higher \(\mathrm{fPAR}\) increases net radiation absorption and induces warming, while lower \(\mathrm{fPAR}\) yields mild cooling (Supplementary \textcolor{blue}{Fig. S2})\\
Collectively, these results indicate that greening effects are nonlinear and depend on how vegetation structure (LAI, EVI) and function (fPAR) co-vary under different moisture and ventilation conditions, providing the basis for the threshold analysis in the next section. 

\subsection*{Climate-dependent thresholds of vegetation driven thermal adaptation}

\begin{figure}[H]
    \centering
    \includegraphics[width=0.9\textwidth]{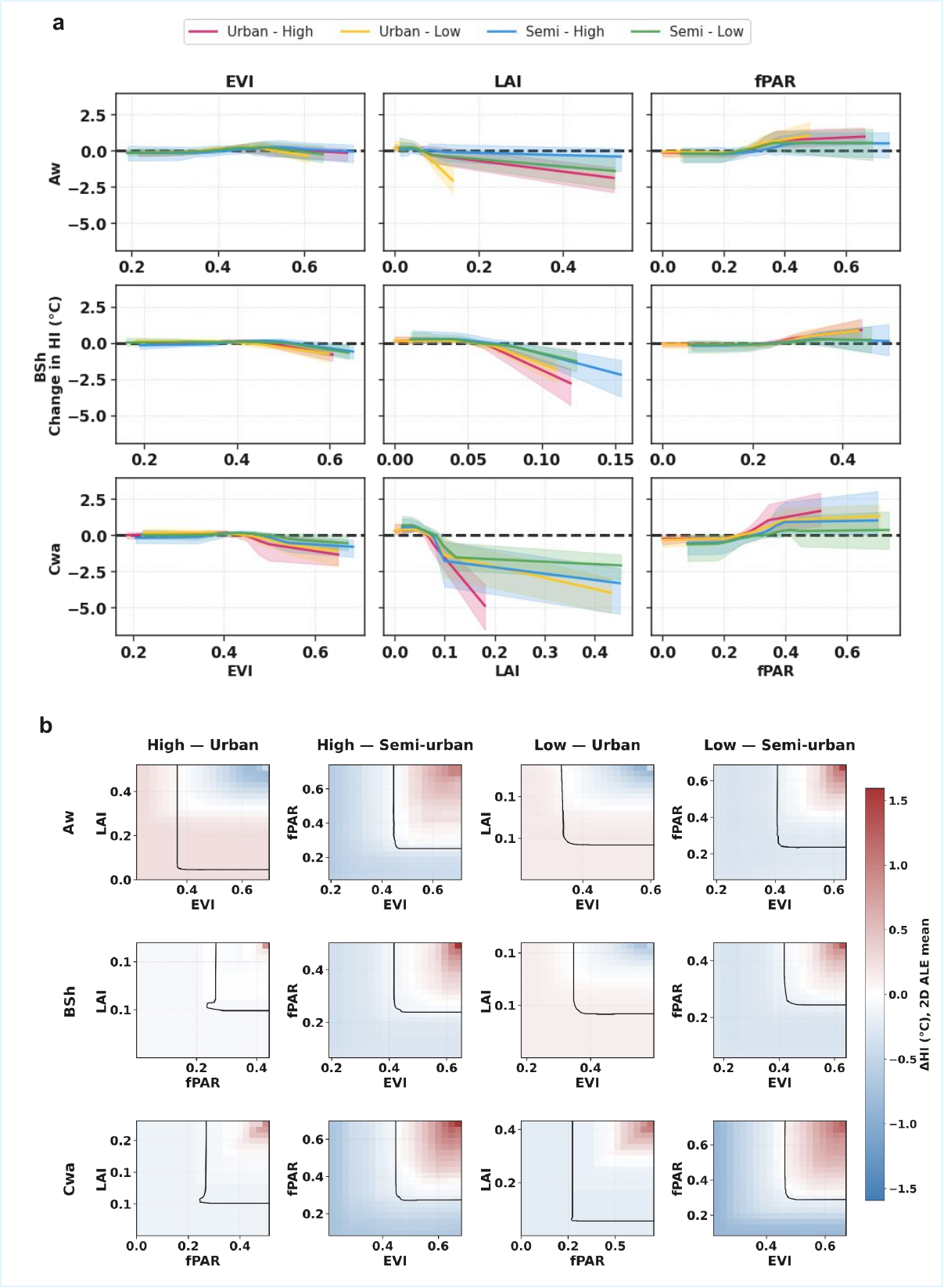}
    \caption{
        \textbf{Greening–productivity interactions and their effects on Heat Index (HI, °C) across climate types and urbanization levels.}
        (\textbf{a}) Partial dependence plots showing the marginal effects of Enhanced Vegetation Index (EVI), Leaf Area Index (LAI), and Fraction of Photosynthetically Active Radiation (fPAR) on HI across urban (high, low) and semi-urban (high, low) classes for tropical (Aw), arid steppe (BSh), and humid-subtropical (Cwa) climates. Shaded bands represent 95\% confidence intervals. 
        (\textbf{b}) Two-dimensional Accumulated Local Effects (2D-ALE) maps showing combined interactions among EVI, LAI, and fPAR pairs, with contours indicating zero-crossing isolines (\(\Delta \mathrm{HI} = 0\)). 
        Red regions denote warming (positive \(\Delta \mathrm{HI}\)) and blue regions denote cooling (negative \(\Delta \mathrm{HI}\)), illustrating nonlinear and synergistic effects of greening and productivity on thermal regulation across urban gradients. All ALE results pool the full 2003–2020 period by design to illustrate “what happens if” we increase or decrease greening features, singly or in pairs, rather than to compare decades.}
    \label{fig:5}
\end{figure}

The preceding analysis showed that vegetation-driven cooling relationships are highly context dependent, indicating climatic thresholds beyond which additional greening may cease to enhance comfort. To characterize these limits, we identified the ranges of vegetation attributes in which their association with HI transitions from negative to positive using the accumulation of local effects (ALE) analysis.(Fig.~\ref{fig:5}). One-dimensional ALE curves describe marginal responses, while two-dimensional surfaces capture joint variations among vegetation metrics.

HI values begin to decline at approximately \(\mathrm{EVI} = 0.4\) across both urban cores and semi-urban rings in cities of varying densities. However, the cooling effect intensifies in humid-subtropical (Cwa) cities when \(\mathrm{EVI} \geq 0.4\), with stronger declines in HI observed within high-density urban cores. In contrast, the magnitude of cooling remains relatively similar between density classes for the tropical (Aw) and semi-arid (BSh) climates. Additionally, EVI values tend to be higher in high-density cities within tropical (Aw) regions but lower in their low-density counterparts, whereas the opposite pattern is evident in Cwa and BSh climates, where low-density cities exhibit relatively higher EVI. (Fig.~\ref{fig:5}a)\\
LAI exhibits a similar pattern to EVI. However, at very low values (\(\mathrm{LAI} \leq 0.05\)), HI tends to increase slightly, indicating mild warming within city environments. The crossover to cooling occurs when \(\mathrm{LAI} \geq 0.05\), after which HI decreases progressively with increasing LAI. The cooling effect of LAI is stronger in high-density cities, both within urban cores and semi-urban rings, particularly in semi-arid (BSh) and humid-subtropical (Cwa) climates. In contrast, fPAR shows an opposite trend: at low values (\(\mathrm{fPAR} \leq 0.25\)), HI decreases modestly, but beyond this range, HI rises sharply, reaching increases of up to \(+2~^{\circ}\mathrm{C}\) at higher fPAR levels in Cwa cities, followed by smaller increases in Aw and BSh cities (Fig.~\ref{fig:5}a). The warming effect of fPAR is consistently greater in urban cores than in their semi-urban fringes, highlighting the influence of built-up density and limited ventilation on canopy-induced humidity buildup.

Building on the 1D ALE results, we use 2D-ALE to map joint responses, highlighting the strongest pairs for each Köppen \(\times\) density group. For tropical (Aw) urban cores across both density classes, the strongest interacting pair identified by the \(H^2\) test was \(\mathrm{EVI}\)--\(\mathrm{LAI}\). An increase in both \(\mathrm{EVI}\) and \(\mathrm{LAI}\) leads to a reduction in the Heat Index (HI), with cooling ranging from \(-0.4\) to \(-1.0~^{\circ}\mathrm{C}\) (Fig.~\ref{fig:5}b), indicating a synergistic effect between greenness and canopy structure. 
In contrast, for the semi-urban fringes across all climate zones, the strongest interaction was observed between \(\mathrm{EVI}\) and \(\mathrm{fPAR}\), where simultaneous increases in both variables (\(\mathrm{EVI} > 0.4\), \(\mathrm{fPAR} > 0.4\)) result in higher HI values, particularly in dense built-up zones, reaching up to \(+1.5~^{\circ}\mathrm{C}\) (Fig.~\ref{fig:5}b). 
Thus, in Aw urban cores, enhancing vegetation structure and greenness (\(\mathrm{LAI}\) and \(\mathrm{EVI}\)) is advisable for achieving a stronger cooling effect, whereas in semi-urban fringes, increasing either \(\mathrm{EVI}\) or \(\mathrm{fPAR}\) individually, rather than jointly, is more effective for mitigating thermal discomfort. 
In humid-subtropical (Cwa) regions, the strongest pair identified was \(\mathrm{LAI}\)--\(\mathrm{fPAR}\) for urban cores and \(\mathrm{EVI}\)--\(\mathrm{fPAR}\) for semi-urban fringes (\ref{fig:5}b). 
Across both cases, cooling dominates at lower vegetation thresholds (\(\mathrm{EVI}<0.3\), \(\mathrm{fPAR}<0.3\), \(\mathrm{LAI}<0.1\)) with HI reductions of \(-0.5\) to \(-1.5~^{\circ}\mathrm{C}\), while joint-high combinations (\(\mathrm{EVI}>0.5\), \(\mathrm{fPAR}>0.5\), \(\mathrm{LAI}>0.2\)) yield a reversal to warming, with HI increases up to \(+1.0\) to \(+1.8~^{\circ}\mathrm{C}\) (Fig.~\ref{fig:5}b). 
These results demonstrate clear thresholds where vegetation-driven cooling transitions to warming under humid and dense urban conditions.

Tree species selection and planting strategies should therefore be guided by these interaction dynamics to optimize local thermal mitigation. The agreement between 1-D and 2-D responses suggests that structural shading (EVI, LAI) remains consistently beneficial, whereas functional greening (fPAR) can shift sign when atmospheric moisture accumulates faster than ventilation disperses it.
Together, these analyses reveal that vegetation–climate interactions exhibit distinct functional regimes, motivating a cross-climate synthesis to understand their physical underpinnings and design implications.

\section*{Discussion}\label{sec3}
Urban greening is central to how cities pursue climate adaptation\cite{li2025global, wong2021greenery}, but its thermal benefits remain uneven across climatic and morphological contexts\cite{li2024cooling}. 
Our results show that vegetation-driven cooling follows distinct climate-dependent regimes that delineate both the potential and the limits of urban thermal adaptation. Structural greening, represented by canopy depth and surface greenness (EVI and LAI), consistently lowers the Heat Index (HI) by about 0.4--1.0~$^{\circ}$C when vegetation density exceeds moderate thresholds (EVI~$\gtrsim$0.3, LAI$\gtrsim$~0.1--0.3). Beyond these ranges, particularly when the Fraction of Absorbed Photosynthetically Active Radiation (fPAR) exceeds $\sim$0.35 in humid or poorly ventilated settings, the cooling effect weakens or reverses, increasing HI by up to 1.0--1.8~$^{\circ}$C. Thus, the adaptation value of greening depends less on vegetation abundance than on how canopy structure and physiological activity interact with background moisture and airflow.

Three physical controls explain these patterns. In semiarid climates (BSh), strong evaporative demand ensures that additional transpiration efficiently lowers HI, while in humid and dense urban cores (Cwa), excess evapotranspiration raises near-surface humidity and dampens sensible heat loss. Compact morphologies further restrict ventilation\cite{chen2017impacts} and promote moisture retention\cite{wang2021urban}, limiting the realized benefits of fPAR. When the canopy structure (EVI, LAI) and moderate physiological activity operate jointly, cooling persists even in humid settings, underscoring the need to balance shading, transpiration, and aerodynamic exchange\cite{lindberg2016impact, rahman2019comparing}.

Translating these insights into design guidance, balanced vegetation structure and function, where canopy depth and greenness enhance shading without excessive evapotranspiration, achieve consistent cooling in arid and tropical climates. In compact and humid environments, the moderation of fPAR along with improved ventilation prevents humidity-induced warming. These findings establish a climate-sensitive framework for harmonizing shading, evapotranspiration, and airflow in urban greening design.

Placed in a broader context, our framework reconciles earlier reports of inconsistent vegetation effects on thermal environments\cite{li2024cooling, yang2024regulation, meili2021vegetation} by identifying the climatic and morphological mechanisms that govern them. Building on global evidence that humidity amplifies urban heat stress\cite{zhang2023increased}, we trace its biophysical origin to the interaction between vegetation physiology and urban form. The resulting comfort-based, humidity-aware paradigm connects canopy processes to perceived heat and unifies previously divergent findings.

Because the burdens of heat and humidity are concentrated in compact, low-ventilation neighbourhoods and among outdoor workers, prioritising climate-sensitive greening in these districts can maximise health benefits and reduce climate inequities. The analytical framework developed here, integrating interpretable machine learning (SHAP--ALE) with physical reasoning, provides a replicable approach for diagnosing complex urban–climate interactions. By quantifying the effects of structural and functional vegetation, it transforms the notion of ``green equals cool'' into a context-sensitive foundation for thermal adaptation.

Although the relationships identified here are derived from statistical associations, the combination of multi-year reanalysis and satellite data captures robust climatological patterns across Indian cities. Finer-scale processes such as canyon airflow, soil-moisture heterogeneity, and species-specific transpiration remain unresolved at 1~km resolution. Future work coupling local microclimate observations with process-based models will strengthen causal interpretation.

Our analysis establishes a mechanistic, data-driven basis for climate-specific greening, revealing how vegetation physiology and urban form jointly constrain the limits of cooling under humid conditions. By linking vegetation function to climate-dependent comfort, this study reframes urban greening as a precision adaptation strategy—transforming it from a symbolic measure into a targeted climate-control tool for resilient cities in a warming world.

\section*{Methods}\label{sec11}
 We selected 138 prominent Indian cities and delineated their core uban boundaries using the IGBP “urban” class from the Terra–Aqua combined MODIS land-cover product (MCD12Q1) \cite{friedl2010modis}, further refined using high-resolution satellite basemaps in ArcGIS \cite{sethi2024urbanization}. For each city, we also defined a semi-urban ring as an equal-area polygon contiguous with the core (Supplementary \textcolor{blue}{Fig. S3}). However, we then extracted 2\,m air temperature and 2\,m relative humidity from the Indian Monsoon Data Assimilation and Analysis (IMDAA) reanalysis \cite{rani2021imdaa} dataset for the entire India and computed the Heat Index (HI) \cite{rothfusz1990heat} from these fields using the NOAA/NWS formulation.
 
 To resolve intra-urban patterns, we downscaled the native 12\,km HI fields to 1\,km resolution using a regression model with physically informed predictors, including land surface temperature (LST) \cite{wan2015myd11a1}, white-sky albedo (WSA) \cite{schaaf2015mcd43a3}, population as a proxy for urban intensity (POP)(\hyperlink{blue}{https://www.worldpop.org/}), net surface radiation (RAD) \cite{munoz2021era5}, 2\,m dew-point temperature (DPT )\cite{rani2021imdaa}, impervious surface fraction (IMP) \cite{huang202130, ren2025improving}, and geospatial covariates (elevation, distance to coast, latitude, land cover). We then (i) ranked features with SHAP (SHapley Additive exPlanations) using a greening/urban-only explanatory model to identify the single most influential driver of HI; (ii) identified the dominant interacting pairs via SHAP interaction analysis and visualized their joint effects on HI; and (iii) quantified the HI response in $^{\circ}\mathrm{C}$ for specified changes in each key feature using 1-D ALE and 2-D ALE, then translated these responses into greening targets needed to achieve a desired reduction in HI. We summarized results by K{\"o}ppen--Geiger climate class and by core versus semi-urban zones, and reported uncertainty via model replicates and block bootstrapping over (city, year). The overall analytical workflow is illustrated in (Supplementary \textcolor{blue}{Fig. S4}).

\subsection*{Study area and Data collection}

We analyzed 138 major Indian cities \cite{sethi2024urbanization}, stratified by Köppen–Geiger climate classification \cite{beck2018present}—tropical savanna (Aw), tropical monsoon (Am), humid subtropical (Cwa), hot semi-arid steppe (BSh), and hot desert (BWh). Within each climate class,  cities were further divided into high- and low-population-density tiers, using the class-specific median as the cutoff. For each city, we delineated a core urban polygon and defined a contiguous equal-area semi-urban ring to enable core–periphery contrasts. For grouped inference, we focused on the best-represented classes—Aw, BSh, and Cwa—which each contain at least 15 cities per density tier; Am and BWh are retained in all model fitting across the full 138-city sample but are excluded from cohort-level summaries due to small counts.

\subsubsection*{In situ data}
In-situ hourly air temperature (T) and relative humidity (RH) measurements were obtained from the HadISD dataset \cite{dunn2016expanding} and utilized to validate 1-km urban Heat Index (HI) predictions. The HadISD dataset encompasses over 9,600 monitoring stations worldwide, with 2,083 in the Global South, and 75 points in India. These T and RH measurements are of high reliability and have been widely used in previous studies \cite{dong2022gsdm,meili2022diurnal,raymond2020emergence}. We first converted the universal coordinated timestamps to indian standard time and then created composites of hourly averages for daytime (14:00–16:00), corresponding to the typical observation times of midday maximum temperature.

\subsubsection*{Satellite data}
We utilized four MODIS products: Land Surface Temperature (LST, MYD11A2; 8-day; $1\,\mathrm{km}$) \cite{wan2015myd11a1}, Enhanced Vegetation Index (EVI, MOD13Q1; 16-day; $1\,\mathrm{km}$) \cite{didan2015mod13q1}, White-Sky Albedo (WSA, MCD43A3; daily; $500\,\mathrm{m}$), and LAI/fPAR (MOD15A2H; 8-day; $500\,\mathrm{m}$) \cite{myneni2002global}. LST and WSA were employed as predictors for downscaling the HI to $1\,\mathrm{km}$ resolution, while EVI, LAI, and fPAR were used to characterize urban greening in the explanatory analyses.

\subsubsection*{Re-analysis data}
We took monthly means of surface net solar radiation (RAD) and $2\,\mathrm{m}$ dew point temperature (DPT) from ERA5-Land monthly aggregated dataset. These variables were resampled from their native $\sim 9\,\mathrm{km}$ resolution to $1\,\mathrm{km}$ to align with the analysis grid. We obtained near-surface air temperature ($2\,\mathrm{m}$) and relative humidity ($2\,\mathrm{m}$) from the IMDAA regional reanalysis for India (NCMRWF/IMD). The 12-hourly IMDAA fields were used to construct daytime heat-stress metrics at the seasonal scale.

\subsubsection*{Demographic data}
We used gridded population (POP; LandScan, $1\,\mathrm{km}$) to stratify cities into density tiers and, where relevant, as a covariate. Cities were assigned to K\"oppen--Geiger classes (Aw, Am, Cwa, BSh, BWh within Indian footprints) for climate-specific grouping. We derived the impervious surface fraction from the Global Impervious Surface Area dataset (GISA) ($30\,\mathrm{m}$) and aggregated it to $1\,\mathrm{km}$ fractional cover. Urban form and land cover were summarized using a global LCZ product ($100\,\mathrm{m}$), which we reclassified into heat-relevant groups, and converted to $1\,\mathrm{km}$ fractional cover. Night-time lights (NTL) data \cite{elvidge2021annual} were obtained from VIIRS DNB monthly composites and resampled to $1\,\mathrm{km}$ as a proxy for built intensity.

Details of all dataset used in this study is described in (Supplementary \textcolor{blue}{Table S1})

\subsection*{Heat Index Estimation}

We constructed India-wide HI from IMDAA temperature and humidity using the NOAA/NWS operational procedure. We converted IMDAA timestamps to IST (UTC$+05{:}30$). For each day, HI was computed at each 12-hourly daytime field, then took it for each pixel and year, the seasonal daytime maximum for March--April--May (MAM). HI values were calculated only where both temperature and relative humidity were valid. We then converted all inputs to the units required by the algorithm and report the final HI in $^\circ\mathrm{C}$ using the NOAA/NWS algorithm as follows:

Let $T_F$ be the near-surface air temperature in $^\circ\mathrm{F}$ and $RH$ the relative humidity in percent ($0$ to $100$). We convert temperature in $^\circ\mathrm{C}$ to $^\circ\mathrm{F}$ by
\[
T_F \;=\; 1.8\,T_{\mathrm{C}} + 32.
\]

\textbf{Simple (Steadman-consistent) Heat Index:}
\begin{align}
HI_{\text{simple}} &= 0.5\!\left(T_F + 61.0 + 1.2\,(T_F - 68.0) + 0.094\,RH\right).
\end{align}

\textbf{Screening Test:}
\begin{align}
HI_{\text{test}} &= \frac{HI_{\text{simple}} + T_F}{2}.
\end{align}
\[
\text{If } HI_{\text{test}} < 80^\circ\mathrm{F}, \text{ we use } HI_{\text{simple}} \text{ and skip the full regression.}
\]

\textbf{Full Rothfusz Regression (when $HI_{\text{test}} \ge 80^\circ\mathrm{F}$):}
\begin{align}
HI_F ={}& -42.379 + 2.04901523\,T_F + 10.14333127\,RH \nonumber\\
& - 0.22475541\,T_F RH - 0.00683783\,T_F^{2} \nonumber\\
& - 0.05481717\,RH^{2} + 0.00122874\,T_F^{2} RH \nonumber\\
& + 0.00085282\,T_F RH^{2} - 0.00000199\,T_F^{2} RH^{2}.
\end{align}

\textbf{Humidity-Range Adjustments:}

\textit{Low humidity adjustment (subtract), if $RH<13\%$ and $80 \le T_F \le 112$:}
\begin{align}
\Delta HI_{\text{low}} &= \left(\frac{13-RH}{4}\right)\,
\sqrt{\frac{17-\lvert T_F - 95\rvert}{17}}, \qquad
\end{align}

\textit{High humidity adjustment (add), if $RH>85\%$ and $80 \le T_F \le 87$:}
\begin{align}
\Delta HI_{\text{high}} &= \left(\frac{RH-85}{10}\right)\left(\frac{87 - T_F}{5}\right), \qquad
\end{align}

\textbf{Convert back to Celsius:}
\begin{align}
HI_{^\circ\mathrm{C}} &= \frac{HI_F - 32}{1.8}.
\end{align}

For each day, we targeted the most thermally stressful period using the daytime 12-hourly IMDAA field(s). Monthly means were then computed for March, April, and May, followed by the formation of a seasonal (MAM) mean HI for each pixel and year over 2003--2020. Missing values were handled conservatively: HI was calculated only where both temperature and humidity data were available, and means were taken over existing records without temporal gap filling. All datasets underwent quality control, were clipped to the India boundary, co-registered to a common $1\,\mathrm{km}$ CRS/grid, and aggregated into annual MAM means. We used bilinear interpolation for continuous rasters and nearest neighbour for categorical layers during resampling. We then normalized predictors per year and season to stabilize learning. When displaying feature values in figures (for example, on ALE or interaction plots), axes were shown in normalized units unless stated otherwise; thresholds and ranges were therefore expressed in normalized space by design.

\subsection*{Ensemble downscaling of Heat Index (HI)}

We built independent, year‐specific models for 2003–2020 to downscale pre-monsoon March–April–May (MAM) Heat Index to a \(1\,\mathrm{km}\) analysis grid using a supervised-learning ensemble. For each year, we assembled a co-registered predictor stack comprising land surface temperature, white-sky albedo, population, surface net solar radiation (converted from \(\mathrm{J\,m^{-2}}\) to \(\mathrm{W\,m^{-2}}\)), distance to coast (km), 2-m dew-point temperature, and impervious surface fraction, with latitude and longitude included to capture broad spatial gradients. All rasters were quality-controlled, clipped to the India boundary, aligned to a common \(1\,\mathrm{km}\) grid, and normalized per year and season. Because MAM marks the pre-monsoon build-up of heat and moisture, we targeted the midday exposure window (\(12{:}00\)–\(16{:}00\) IST) when outdoor stress peaks. The target variable for model training was the \(1\,\mathrm{km}\) MAM mean HI described earlier, derived by applying the NOAA/NWS Heat Index formulation to IMDAA daytime fields and aggregating from daily to monthly and then to seasonal scales on the same grid.

Unless otherwise noted, we used all valid pixels, then created a random \(70{:}30\) train–test split with a fixed seed \((7)\), yielding a large, geographically diverse training set each year and a held-out test set for year-specific generalization. Two complementary learners were fitted: a Random Forest regressor for the mean structure \((n_{\mathrm{estimators}}=150,\ \mathrm{max\_features}=\mathrm{sqrt},\ \mathrm{no\ depth\ cap},\ n_{\mathrm{jobs}}=-1)\) providing robust, low-variance predictions across the bulk of the distribution; and a Gradient Boosting regressor with quantile loss at \(\tau=0.90\) \((\mathrm{alpha}=0.9,\ \mathrm{learning\_rate}=0.05,\ \mathrm{max\_depth}=5,\ \mathrm{up\ to}\ 1000\ \mathrm{estimators\ with\ early\ stopping},\ \mathrm{validation\_fraction}=0.1)\) designed to capture the high-HI tail. We selected these settings after pilot hyperparameter sweeps on representative years using the same train–test protocol, optimizing mean absolute error on the held-out data while checking tail bias. Model robustness was verified by refitting with perturbed hyperparameters around these values, which yielded materially similar predictive skill and spatial patterns. To reduce class imbalance for the quantile learner, we oversampled training pixels above the empirical 90th-percentile threshold so that extremes compose about \(30\%\) of its training set, using replacement only when needed for the extreme subset. We combine predictions by a pixelwise maximum,
\[
\hat{y}_{\mathrm{ens}}(i)=\max\!\bigl(\hat{y}_{\mathrm{RF}}(i),\,\hat{y}_{\mathrm{Q90}}(i)\bigr),
\]
which explicitly prioritizes extreme-heat detection and mitigates the common negative bias of mean-loss learners in the upper tail. In practice, the Random Forest governs typical conditions, while the quantile model elevates predictions where the learned 90th percentile exceeds the Random Forest estimate. We evaluated the Random Forest, the quantile model, and their max-ensemble on the held-out test set each year using mean absolute error in \(^{\circ}\mathrm{C}\), and we proceed with the ensemble for mapping.

Our design choices reflect four safeguards that matter for risk-focused urban heat applications. First, to capture spatial signals beyond the physical predictors, we included latitude, longitude, and a GEO basis so the models can absorb broad gradients related to climate zones, elevation, and regional context. Second, to be extreme-aware, we used quantile loss at \(\tau=0.90\) together with tail oversampling, aligning the learning objective with operational concern for high HI. Third, to maintain robustness, we chosen Random Forest hyperparameters that offer a strong bias–variance trade-off and we curbed overfitting in the boosting stage with early stopping. Fourth, to ensure reproducibility and stable training, we fixed the random seed and normalize inputs per year and season. Together, these steps produce year-wise \(1\,\mathrm{km}\) HI fields that retain fidelity for typical conditions while minimizing underestimation of dangerous high-HI events, which is essential for analysis and planning.

After aligning station--year pairs, we evaluated model performance using five metrics: Pearson correlation ($r$), mean absolute error (MAE), root-mean-square error (RMSE), mean bias (model $-$ obs), and coefficient of determination ($R^2$). Agreement between observed and modeled HI values is summarized in a scatter plot with a 1:1 reference line (Supplementary\textcolor{blue}{Fig. S5}). The validation encompassed a total of  $=1032$ paired station–years, yielding $r=0.856$; MAE $=1.28~^\circ$C; RMSE $=1.72~^\circ$C; Bias $=-0.23~^\circ$C; $R^2=0.709$. The regression cloud clusters tightly around the 1:1 line, indicating strong model–observation agreement, with a slight cool bias ($-0.23~^\circ$C) and residual spread consistent with sub-grid variability and station-exposure differences.

\subsection*{Explainable machine learning for urban greening impacts}
We quantified how greening and urban form shape thermal discomfort by explaining the modelled Heat Index with SHAP and by tracing responses with Accumulated Local Effects. SHAP (Shapley Additive Explanations) attributes, in $^\circ$C, the contribution of each feature to deviations of the prediction from a common baseline at each pixel, while ALE shed light on how the prediction changes as a feature varies over its empirical range after accounting for correlations. We used SHAP for local, signed, additive attributions that aggregate cleanly to cities, density cohorts, and climate classes. ALE was used to identify effect sizes and thresholds that are robust to feature interdependence.

To avoid explaining Heat Index with its own drivers, we excluded variables used in the downscaling target construction or their close surrogates. The explanatory model therefore uses only independent, biophysical and urban indicators: EVI, LAI, and fPAR to represent surface greenness, canopy structure, and photosynthetic activity; night-time lights as a proxy for built intensity; and Local Climate Zone groups as land-cover and morphology controls. This separation prevents leakage between stages and keeps the interpretation focused on vegetation and form rather than on reanalysis temperature or humidity.

\subsubsection*{Shapley Additive Explanations (SHAP)}
SHAP belongs to the class of additive feature-attribution methods grounded in cooperative game theory, and it is the unique solution that satisfies local accuracy, missingness, and consistency; this gives us stable, comparable contributions that sum to the model output \cite{lundberg2017unified}. We adopted the SHAP framework because it unifies prior explanation approaches under single theoretical foundation and provides algorithms suitable for our tree-ensemble explainer model, allowing fast local explanations and clean aggregation to city, cohort, and climate classes \cite{lundberg2017unified}. SHAP, or Shapley Additive Explanations, assigns each feature a signed contribution to a model’s prediction in the same units as the target. It gives local explanations at the pixel or instance level, and these can be summed or averaged to neighborhoods, cities, or cohorts for policy use. It is widely used in climate and heat-risk modeling to identify dominant drivers, compare importance across contexts, and pair global rankings with per-instance insight \cite{yuan2022natural, kim2022explainable}. We computed SHAP values in three steps. First, we built a national background distribution of features by sampling across cities and years and compress it with MiniBatch K-means to a small set of representative centroids. This background fixes a stable expected value that anchors all attributions. Second, we trained a single Random Forest on the pooled city–year data with only the greening and urban predictors listed above and we recorded the model base value and out-of-bag skill for provenance. Third, for each city and year, we predicted Heat Index on valid $1\,\mathrm{km}$ pixels inside the city polygon, compute TreeSHAP values in chunks against the fixed background, and save per-pixel attributions along with predictions and the baseline. We then summarized SHAP to city and cohort scales, form ranked importances from mean absolute SHAP, and build dependence plots that show how the sign and magnitude of a feature’s attribution vary over observed feature values. 

\subsubsection*{Accumulated Local Effects (ALE)}
We complemented SHAP with ALE in one and two dimensions. ALE, or Accumulated Local Effects, shows how the model’s prediction changes on average as a feature varies over its observed range while accounting for correlations with other features. In 1D it yields a centered curve in the same units as the target, and in 2D it maps joint regions where pairs of features tend to cool or warm \cite{yang2024mitigating, robertson2023gut}. For ALE-1D, we partitioned each feature’s empirical range into quantile bins, estimated the local derivatives from neighbours, and integrated them to obtain a centered curve $\Delta \mathrm{HI}(x)$ that reads as “how HI changes when this feature increases,” with units of $^\circ$C and zero at the typical state. These curves provided effect sizes and onset ranges that are less biased by correlated predictors than partial dependence. 

For ALE-2D, we took grid pairs of features and computeed a centered surface $\Delta \mathrm{HI}(x_1,x_2)$ that reveals joint regimes where combinations cool or warm. We selected the strongest pairs per climate–density cohort using an interaction strength statistic and visualized the corresponding surfaces. We identify the strongest pair per Köppen \(\times\) density group using an \(H^2\) interaction test \cite{friedman2008predictive}. A pair is selected if its last five year mean \(H^2\) is both high and stable, that is, within one standard deviation (or 0.02 at minimum) of its long run 18 year mean.Otherwise, we fall back to the long run leader (Supplementary \textcolor{blue}{Fig. S6}). This guards against short term noise and still recognizes genuine recent shifts. Together, SHAP highlights which drivers the model relies on and in which direction, while ALE quantifies how much HI changes for specified changes in each key feature and for the top feature pairs, enabling threshold and design inference without relying on correlation alone.

\subsubsection*{Statistical analyses}
To assess the statistical distinctiveness of feature contributions to the Heat Index (HI) across climatic and urbanization contexts, we performed pairwise Kolmogorov--Smirnov (KS) tests\cite{massey1951kolmogorov} on the distributions of SHapley Additive exPlanations (SHAP) values.

For each climate type---tropical (Aw), semi-arid (BSh), and humid-subtropical (Cwa)---and for each urbanization stratum (Urban-core and Semi-urban ring) under both high- and low-density categories, SHAP values were computed using the trained machine learning model for the period 2003--2020. 
To ensure comparability, we aggregated annual mean SHAP values per city, yielding one representative distribution per climate--density--urbanity group (Aw-H-U, Aw-H-Su, Aw-L-U, Aw-L-Su, etc.). Each distribution captures the direction and magnitude of the predictor’s contribution to HI under distinct environmental and morphological conditions.

We applied the two-sample Kolmogorov--Smirnov (KS) test to all possible group pairs for each predictor. The KS test is a non-parametric method used to evaluate whether two samples originate from the same underlying distribution. For each pairwise comparison, we report the KS statistic (\(D\)) and the associated p-value, marking significance thresholds at $p~<~0.10 (*)$, $p~<~0.05 (**)$, and $p~<~0.01 (***)$. This analysis allows direct quantification of how feature importance distributions (SHAP values) vary across climatic and urban morphological regimes. The results are reported at (Supplementary \textcolor{blue}{Fig. S7}) 

We also reported two standardized effect metrics :\(r_{\mathrm{SD}}\) expresses a feature’s effect size as a fraction of the response variability (standard deviation), and  \(r_{\mathrm{RMSE}}\) expresses effect size relative to predictive error Across climate–density cohorts, greening features show \(r_{\mathrm{SD}}=0.10\text{–}0.35\) (Supplementary \textcolor{blue}{Fig. S8})and \(r_{\mathrm{RMSE}}=0.3\text{–}1.0\) (Supplementary \textcolor{blue}{Fig. S9}), indicating moderate to strong explanatory power and effects that are practically meaningful relative to predictive error. In contrast, NTL is consistently small, with most \(r_{\mathrm{RMSE}}<0.3\)  \cite{jacobs2018use}.

\backmatter

\bmhead{Acknowledgements}

This research was primarily funded by IIT Gandhinagar. The authors also thank the members of the Machine Intelligence and Resilience Laboratory at IIT Gandhinagar for their valuable discussions and constructive feedback on this manuscript.

\bmhead{Competing interests}

The authors declare that they have no competing interests.\\
\bmhead{Data and code availability} 
The codes used for the analyses done in this study are avialbale in the github repository, \href{https://github.com/angana16/Heat-Index-and-urban-vegetation/upload/main}{https://github.com/angana16/Heat-Index-and-urban-vegetation/upload/main}. All datasets used in this study are publicly available and are properly referenced at the point of first mention in the Methods section within the manuscript. \\
\bmhead{Author contribution}: Conceptualization: A.B., A.D., A.S.K, R.D., U.B.;
Design of framework: A.B., A.S.K, A.D., U.B.; Methodology: A.B, A.S.K., A.D.; 
Data curation: A.S.K., A.B.; 
Formal analysis: A.B., A.D., A.S.K.; 
Validation: A.B, A.S.K.; 
Visualization: A.D., A.B., A.S.K.; 
Writing-original draft: A.B., A.D., A.S.K, U.B.; 
Writing-review-editing: A.B., A.S.K, A.D., R.D., U.B..





\end{document}